# Towards Combining HTN Planning and Geometric Task Planning

Lavindra de Silva     Amit Kumar Pandey     Mamoun Gharbi     Rachid Alami

*Abstract*—In this paper we present an interface between a symbolic planner and a geometric task planner, which is different to a standard trajectory planner in that the former is able to perform geometric reasoning on abstract entities—tasks. We believe that this approach facilitates a more principled interface to symbolic planning, while also leaving more room for the geometric planner to make independent decisions. We show how the two planners could be interfaced, and how their planning and backtracking could be interleaved. We also provide insights for a methodology for using the combined system, and experimental results to use as a benchmark with future extensions to both the combined system, as well as to the geometric task planner.

## I. INTRODUCTION

The past few years have seen a great deal of interest in interfacing symbolic and geometric reasoning. A common theme has been to define how geometric entities and capabilities should be meaningfully used within symbolic planning, and how symbolic information can be used, perhaps as heuristics, in geometric planning. In this paper we follow this trend. Broadly, we are interested in the link between a geometric task/action and a symbolic action, what geometric information we should publicise to the symbolic (resp. geometric) planner, and how can we include them in symbolic (resp. geometric) states and actions. An unavoidable issue that arises when combining the two planning approaches is "backtracking"—trying alternative options for choices that were already made during planning. Both approaches are capable of backtracking at their own levels when a plan being pursued turns out to not work: the symbolic planner when some precondition is not met, and the geometric planner when a path/trajectory being planned cannot avoid a collision. It is therefore important to decide how we can effectively interleave geometric and symbolic backtracking [1], [2], and how we should switch between them.

In [2], [3] the authors discuss algorithms for geometric backtracking by extending the Justin robot with symbolic-geometric planning capabilities: they use the JSHOP2 [4] HTN planner for the symbolic planning component and a specialised path planner for the geometric one. Unlike our approach, the authors keep the symbolic state orthogonal to the geometric state: changing, for instance, the pose of an object on a table has no consequence on the symbolic state. We require symbolic and geometric states to be intertwined, which is natural in some domains. To this end, we derive symbolic facts from the geometric state and use these in symbolic planning.

To interface symbolic and geometric planning, the authors in [5], [6] introduce "semantic attachments," which associate selected predicates in the planning domain to external procedures called at runtime to evaluate the predicates. In [7], the semantic attachments are implemented using a trajectory planner that computes collision free trajectories; if one exists, then the corresponding semantic attachment evaluates to *true*, and *false* otherwise. Likewise, we use "evaluable predicates," with a minor difference where we link them to more abstract entities called geometric *tasks*, which may or may not invoke a trajectory planner. The authors also introduce "effect applicators" in effects of actions that consult the geometric planner to set certain state variables (e.g. position and orientation of a moved object) in the symbolic domain. Effect applicators, however, cannot make decisions between different outcomes, such as choosing where to place an object. In our work we do want to give the geometric planner some leeway to make such choices.

In some works there is a tight integration between symbolic and geometric planning. In [8], for instance, a hierarchical planner plans all the way down to the level of geometric actions. Similarly, [9] describes a special purpose hierarchical planner combined with a geometric motion planner for planning, and then executing the most basic actions while the plan is still being constructed. This is different to the work described above which formulate a complete plan first, before executing it. The Asymov [10] system is a combined task and motion planner for problems that are difficult to solve when the symbolic planner is in control of the geometric search, e.g. in the geometric Towers of Hanoi problem that the authors present. Compared to other approaches, in Asymov the geometric planner uses the symbolic planner—as well as the symbolic model of the domain—as a heuristic when choosing roadmaps during geometric search. Similarly, in [11] a symbolic planner guides a sampling-based motion planner, which in turn sends back utility estimates to improve the guide in the next iteration.

Unlike previous approaches, our work concerns the use of a *geometric task planner* instead of a typical trajectory planner; the former lets us define the interface to symbolic planning in a more meaningful way—by providing a higher level of abstraction to low level geometric actions like picking and placing—and also gives more leeway to the geometric level to make decisions. Unlike past work we are also interested here in a principled methodology for developing symbolic-geometric planning domains: this paper provides useful insights in this direction. We present an initial prototype implementation of the combined planning framework including basic interleaved backtracking, and an analysis of its runtime performance;

*This work was conducted within the EU SAPHARI project (www.saphari.eu) funded by the E.C. division FP7-IST under contract ICT-287513.
**LAAS/CNRS, 7, Av. du Colonel Roche, 31077 Toulouse, France.
{ldesilva,akpandey,magharbi,rachid}@laas.fr

we intend to use these results as a benchmark for future experiments with different backtracking strategies currently being developed, and to develop better heuristics for the geometric task planning component itself.

## II. BACKGROUND

In this paper we refer to the popular STRIPS classical planning language [12]. More importantly, we make use of the Hierarchical Task Network (HTN) planning formalism. While classical planners focus on bringing about states of affairs or "goals-to-be," HTN planners focus on solving *abstract tasks* or "goals-to-do." In this paper we use a popular type of HTN planning called "totally-ordered" HTN (henceforth simply referred to as HTN) planning, which has proven to be efficient in certain domains [13]. An HTN *planning problem* is the 3-tuple $\langle d, S_0, \mathcal{D} \rangle$, where $d$ is the sequence of (primitive or abstract) tasks to solve, $S_0$ is an initial state as in classical planning, and $\mathcal{D}$ is an HTN *planning domain*. Specifically, an HTN *planning domain* is the pair $\mathcal{D} = \langle \mathcal{A}, \mathcal{M} \rangle$ where $\mathcal{A}$—the primitives of the domain—is a finite set of operators as before, and $\mathcal{M}$ is a finite set of *methods*. A *method* is a tuple consisting of the name of the method, the abstract task that the method is used to solve, a precondition specifying when the method is applicable (like an operator's precondition), and a *body* indicating which tasks are used to solve the task associated with the method. The method-body is a sequence of primitive and/or abstract tasks. The planning process works by selecting applicable reduction methods from $\mathcal{M}$ and applying them to abstract tasks in $d$ in a depth-first manner. In each iteration this will typically result in $d$ becoming a "more primitive" sequence of tasks. The process continues until $d$ has only primitive tasks left. At any stage in the planning process, if no applicable method can be found for an abstract task, the planner essentially "backtracks" and tries an alternative method for an abstract task refined earlier.

For the geometric counterpart, we adopt the approach of finding a solution in a discrete space of candidate grasps and placements [3], [14] for tasks involving picking and placing. Basically, our geometric task planner (GTP)[1] iterates in a four dimensional search space, consisting of a set of *agent "effort" levels*, a set of *discrete grasps*, a set of *object placement positions*, and a set of *object placement orientations* (see Figure 1). For each object, sets of possible grasps are pre-computed and stored for the anthropomorphic hands and the robot's gripper, which are later filtered based on task requirements and the environment. The amount of "effort units" required to perform certain tasks like moving the head, extending an arm, and standing up are predefined; to this end, we have made simplifying assumptions about which tasks (e.g. head movements) require less effort than others (e.g. standing up). At runtime, sets of placement positions and possible orientations of objects are dynamically extracted, based on the environment, the task, and restrictions on how much effort should be put into the task. These sets are then weighted based on the environment and situation, with criteria such as grasp and placement stability, feasibility of simultaneous grasps by two agents, the agent's visibility of the object, and estimated effort to see and reach it.

[1]We also use GTP as an abbreviation for geometric task planning.

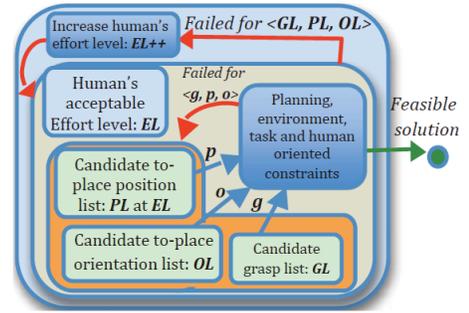

Fig. 1: An overview of the geometric task planner (GTP).

The advantage of the GTP framework is that a variety of day-to-day tasks like *showing*, *giving*, *hiding*, and *making accessible* can be represented in terms of different constraints, based on factors like reachability, visibility, desired effort level, and the ability to grasp. A geometric solution is found using a constraint hierarchy based approach, by carefully introducing these constraints successively at different stages of planning. This facilitates the reduction of the search space successively, before introducing relatively more computationally expensive constraints.

We shall now briefly highlight the GTP algorithm. The outermost loop starts from the lowest estimated effort required (to view and reach objects) for the task and incrementally moves to the highest. For each such effort estimate, the algorithm iterates on the candidate points where the task could be performed, excluding those that did not work with lower effort estimates. For each such point, the algorithm iterates on the possible object grasps if it is not already in the gripper, excluding those that already failed for lower effort estimates. For each such grasp, if a collision-free pick is possible—i.e. there is a path to a configuration associated with the grasp—the algorithm tries the different possible object orientations, excluding those that are in collision (e.g. with a surface), and those that do not satisfy the visibility threshold and other task oriented symbolic constraints, such as maintaining the object facing front. Finally, the planner obtains a tuple consisting of a grasp, a placement position and a placement orientation, which is then used to find a collision free trajectory.

## III. A SYMBOLIC-GEOMETRIC PLANNING EXAMPLE

In this section we detail a concrete domain that illustrates how symbolic and geometric planning is interfaced and combined. We also highlight our approach to interleaving HTN and GTP backtracking, but leave the detailed algorithms for a separate paper. Suppose a PR2 robot is working as a library receptionist. Library members reserve books online with their membership ID, which is also used to top up their library credit. The ID can be used to look up membership details like email address and books reserved and borrowed. Reserved books are collected in person from the library. Once an online reservation is made, (human) librarians make the books accessible (i.e. *reachable* without navigating from the current position, and *visible*) to the PR2 on an adjacent table.

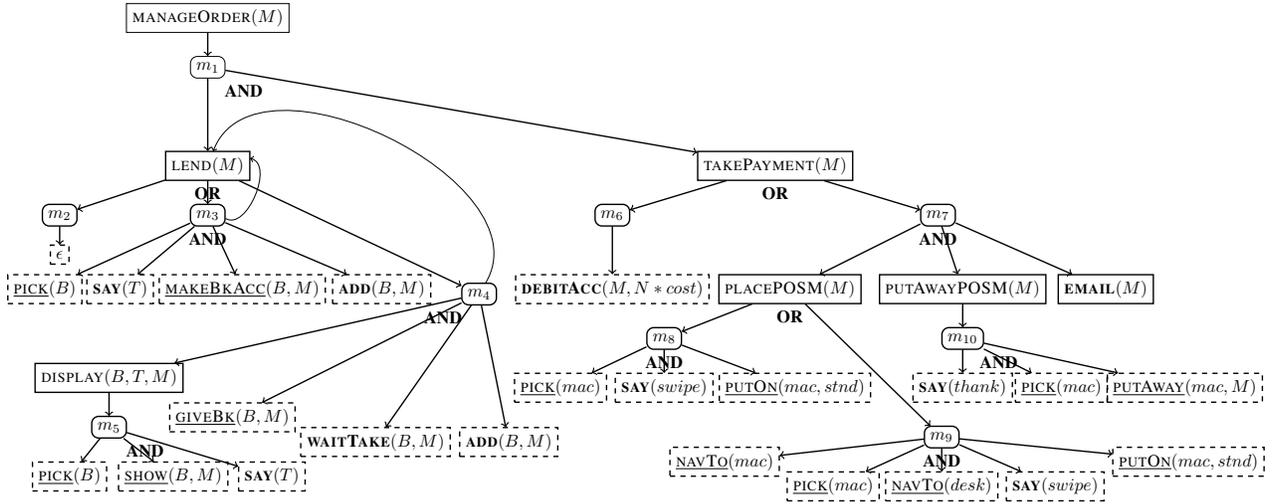

Fig. 2: The part of the HTN domain that handles online book reservations. Solid rectangles are HTN abstract tasks, rounded rectangles are methods (where the incoming vertex is the task that the method solves, and outgoing vertices are the tasks in the method's body), and dashed rectangles are actions. Standard HTN actions are in bold and GS actions are underlined.

| Action/HTN Task | M | Precondition | Method-body/Action-effects |
|---|---|---|---|
| MANAGEORDER($M$) | $m_1$ | $held(B, M)$ | LEND($M$) · TAKEPAYMENT($M$) |
| LEND($M$) | $m_2$ | $\forall B, \neg held(B, M)$ | $\epsilon$ |
| | $m_3$ | $held(B, M) \wedge title(B, T)$ | PICK($B$) · SAY($T$) · MAKEBKACC($B, M$) · ADD($B, M$) · LEND($M$) |
| | $m_4$ | $held(B, M) \wedge title(B, T) \wedge \neg hvy(B)$ | DISPLAY($B, T, M$) · GIVEBK($B, M$) · WAITTAKE($B, M$) · ADD($B, M$) · LEND($M$) |
| DISPLAY($B, T, M$) | $m_5$ | true | PICK($B$) · SHOW($B, M$) · SAY($T$) |
| TAKEPAYMENT($M$) | $m_6$ | $numLent(M, N)$ | DEBITACC($M, N * cost$) |
| | $m_7$ | $numLent(M, N) \wedge cred(M, C) \wedge (C < N * cost)$ | PLACEPOSM() · PUTAWAYPOSM($M$) · EMAIL($M$) |
| PLACEPOSM() | $m_8$ | $reachable(mac, pr2)$ | PICK($mac$) · SAY($swipe$) · PUTON($mac, stnd$) |
| | $m_9$ | true | NAVTO($mac$) · PICK($mac$) · NAVTO($desk$) · SAY($swipe$) · PUTON($mac, stnd$) |
| PUTAWAYPOSM($M$) | $m_{10}$ | true | SAY($thank$) · PICK($mac$) · PUTAWAY($mac, M$) |
| SAY($T$) | | true | $\{spoke(T)\}$ |
| MAKEBKACC($B, M$) | | $held(B, M) \wedge makeAcc(B, M)^?$ | $\{\neg held(B, M), lent(B, M)\}, makeAcc(B, M)^-, makeAcc(B, M)^+$ |
| WAITTAKE($B, M$) | | $held(B, M) \wedge gave(B, M)$ | $\{\neg held(B, M) \wedge \neg gave(B, M) \wedge lent(B, M)\}$ |
| ADD($B, M$) | | $lent(B, M) \wedge numLent(M, N)$ | $\{\neg numLent(M, N), numLent(M, N + 1)\}$ |
| GIVEBK($B, M$) | | $held(B, M) \wedge give(B, M)^? \wedge \neg hvy(B)$ | $\{gave(B, M)\}, give(B, M)^-, give(B, M)^+$ |
| PICK($O$) | | $pick(O)^?$ | $pick(O)^-, pick(O)^+$ |
| SHOW($O, M$) | | $show(O, M)^?$ | $show(O, M)^-, show(O, M)^+$ |
| DEBITACC($M, Cost$) | | $cred(M, C) \wedge (C \geq Cost)$ | $\{\neg cred(M, C), cred(M, C - Cost)\}$ |
| EMAIL($M$) | | $lent(B, M)$ | $\{emailed(M)\}$ |
| PUTON($O_1, O_2$) | | $putOn(O_1, O_2)^?$ | $putOn(O_1, O_2)^-, putOn(O_1, O_2)^+$ |
| PUTAWAY($O, M$) | | $putAway(O, M)^?$ | $putAway(O, M)^-, putAway(O, M)^+$ |
| NAVTO($Obj$) | | $navTo(Obj)^?$ | $navTo(Obj)^-, navTo(Obj)^+$ |

TABLE I: The table for Figure 2. The top half (above the thick line) are methods (M) and the bottom half are operators. For legibility, empty sets have been ommitted from the table.

The HTN domain is illustrated graphically in Figure 2 and detailed in Table I. The top-level HTN task is MANAGEORDER($M$) for member $M$. It has one method (named $m_1$) with two subtasks: LEND($M$) and TAKEPAYMENT($M$). The first is associated with three methods—$m_2, m_3, m_4$—which are tried in that order. Method $m_2$ trivially succeeds if there are no (more) books held by the member and hence no more books to give. If it is not applicable, $m_3$ is tried, which has the following actions: pick (via GTP) from the adjacent shelf a book reserved by the member, speak out the title, make it accessible to the member on the desk, perform some bookkeeping—e.g. send the current total to the Point-of-Sale (POS) machine—and then recursively call LEND($M$).

Method $m_4$ starts with an HTN abstract task to display a book, which refines into the three steps focussed on showing a book to the person. The abstract task is followed by giving the book to the person,[2] waiting for it to be taken—which relies on the gripper angle and force sensors to check if the book has successfully been taken—and then the bookkeeping action as before, ending with a recursive call to LEND($M$). Note that a book is given only if it is deemed light enough to be directly taken from the gripper, and that by forcing an ordering on $m_3$ and $m_4$ we are encoding a preference for placing a book on the table over handing it (directly) to the member, allowing the member to pick up the book and put it in a bag/handbag at his/her own pace.

The TAKEPAYMENT($M$) task of $m_1$ has methods $m_6$ and $m_7$. Method $m_6$ has a single action to debit the account corresponding to the member ID according to the number of books lent (all books have the same cost), if there is enough credit $C$ in the member's account. If not, the member must pay by credit card ($m_7$). The PLACEPOSM($M$) task refines into two methods for giving the POS machine: if the machine is (likely) reachable it is simply picked up, but if not—presumably because the (shared) machine is with the receptionist at an adjacent desk—the PR2 navigates to the machine, picks it up, and navigates back; the PR2 then asks the user to swipe the card and enter the PIN, and then puts the machine on the POS machine stand. The PUTAWAYPOSM($M$) task includes actions to thank the person and to put the machine somewhere that is away from the person's reach and visibility, and EMAIL($M$) emails the member an invoice.

---

[2]We could also imagine more generic GIVEBK($B, M$) and MAKEBKACC($B, M$) actions that can handle any object type or give and make books accessible for reasons other than lending.

This HTN domain serves to highlight some key features. First, the HTN developer interfaces with the GTP using evaluable predicates. To this end, every (relevant) GTP task $t$ is associated with an evaluable predicate—denoted $t^?$—in the HTN domain (e.g. GTP task SHOW($O, H$) has evaluable predicate $show(O, H)^?$ for some object $O$ and human $H$), which evaluates to $true$ if $t$ has a GTP solution and $false$ otherwise. Whenever such an evaluable predicate is mentioned in the precondition of an operator (resp. action), we call it a *Geometric-Symbolic* (GS) operator (resp. action). A GTP task $t$ is also associated with an add list function—denoted $t^+$—and a delete list function—denoted $t^-$—which are the (possibly empty) add and delete lists for $t$ computed by the GTP, based basically on the world resulting from the solution that was found for $t$ on calling evaluable predicate $t^?$. Add list function $show(o_1, h_1)^+$ might, for instance, return the set $\{visible(o_1, h_1), accessible(o_1, h_1)\}$ and $show(o_1, h_1)^-$ the set $\{visible(o_3, h_1), accessible(o_3, h_1)\}$: that is, after making object $o_1$ visible to human $h_1$, the object is also accessible to $h_1$, but object $o_3$ is no longer visible nor accessible to $h_1$. The effects of a GS operator is the combination of its "static" add and delete lists with those obtained via the add and delete list functions.

An important concept we exploit is that of a "shared predicate" (or "shared literal"), which is a standard literal that is based on geometrical properties and hence modelled more accurately by the GTP. For example, predicate $reachable(o, h)$ in the HTN domain (see method $m_8$), which specifies that object $o$ is reachable to human $h$, is derived from the 3D world with a heuristic based on the area covered on extending the robot's arm with respect to all its degrees of freedom [1].

We highlight some interesting interleaving of GTP and HTN planning—backtracking in particular—possible with the domain described. Suppose the library reception desk is small and somewhat cluttered, and that a member has reserved two big books $b_1$ and $b_2$. Assume there is enough space on the table to make one of them accessible (to the member), but not enough space to make them both accessible, nor to make one accessible but give the other (directly), as the books are so big that they block the robot-arm's path to the person. Figure 3 shows a part of a possible combined HTN-GTP planning scenario. In the figure, $b_1$ is successfully picked and made accessible—i.e. the GTP tasks (third column) corresponding to the PICK($b_1$) and MAKEBKACC($b_1, m$) GS actions (second column) are successfully planned. Then, LEND($M$) is recur-sively called during HTN planning. However, according to our scenario, the attempt to make book $b_2$ accessible (after picking it up) will fail. At this point the GTP will backtrack all the way up to PICK($b_1$) (third column) but not find a way to reposition the first book so as to make the second accessible. The system will then resort to HTN bactracking, which will choose the alternative method $m_4$ to give $b_2$ directly to the person.[3] According to our initial scenario, this will also fail even after the GTP backtracks to reposition $b_1$ (not shown in the figure). The HTN planner will then backtrack once again up the hierarchy and perhaps choose method $m_4$ to give $b_1$ directly to the person, after which it should be possible to make $b_2$ accessible and continue planning.

The last column of Figure 3 shows the mapping of the four "compound" GTP tasks in the third column into grasp and placement actions. A more interesting domain would be where these actions are encapsulated into one compound PICKMAKEACCESSIBLE($O, H$) task (for object $O$ and human $H$), instead of the two tasks PICK($O$) and MAKEACC($O, H$). This will allow the GTP to backtrack from PLACE($O$) to GRASP($O$), allowing failure to be detected early in HTN planning—when PICKMAKEACCESSIBLE($O, H$) fails because a (future) GTP task/action (PLACE($O$) in our example), due to be tried later by the HTN planner, is predicted by the GTP to be impossible. Such encapsulation can also minimise GTP backtracking by, from the outset, planning with respect to definite future GTP actions (e.g. planning the grasp with respect to the placement).

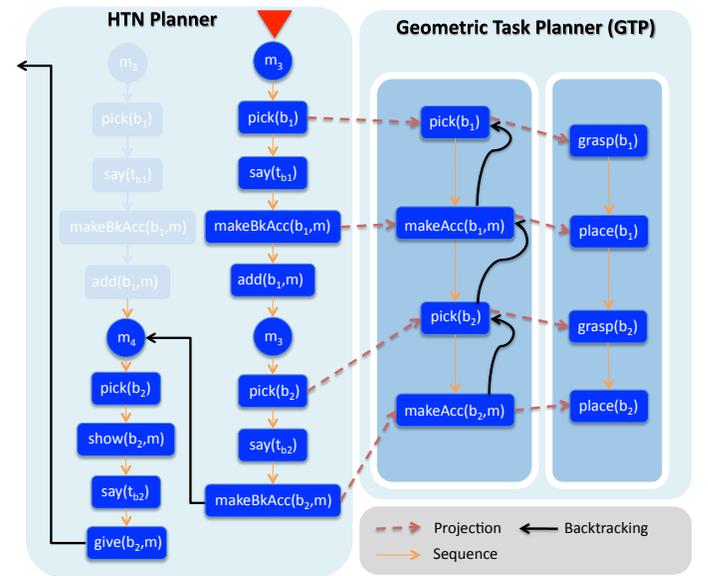

Fig. 3: Combined backtracking scenario for Figure 2.

## IV. IMPLEMENTATION

As mentioned in Section II, we have adapted an HTN planner and implemented a GTP. We have also sufficiently implemented the algorithms presented in this paper to gain some valuable insights. In our setup, the (real) PR2 is in a room with objects like tables, shelves, and chairs, and the same is modelled in 3D, onto which humans and objects (e.g. books) are dynamically projected whenever they are detected in the real room by respectively a Kinect (Microsoft) sensor, and PR2's stereo cameras coupled with a pattern-based marker detection module. For 3D visualization and planning we use the Move3D software [15].

---
[3]We assume that the book is light enough to be handed to the person.

Our current GTP implementation has some of the functionality depicted in Figure 3, including: *(i)* storing the sequence of GTP tasks pursued so far; *(ii)* basic backtracking to find an alternative solution for a chosen GTP task in the sequence; and *(iii)* computing predicates such as $visible(O, A)$, $reachable(O, A)$, $on(O, O_2)$, $inside(O, O_2)$ and $coveredBy(O, O_2)$ for an agent $A$ and objects $O, O_2$, where the first two are computed based on the concept of *mightability maps* presented in [1]—which the authors showed to be computed and updated fast enough for online HRI—and the other facts using techniques from the geometric analysis of the 3D world model, and domain specific heuristics.[4]

We demonstrate our implementation in figures 4 and 5, where the second is essentially a screen dump of the $visible(O, H)$ and $reachable(O, H)$ shared literals computed. Figure 4 (a) is the initial state with two books on the table next to the PR2, and a small white platform in front of it to exchange objects. Planning starts with the PR2 picking (b) the grey book and making it accessible (c) to the human on the platform. The same is done for the white book in (d) and (e). The position of the white book, however, makes the grey one no longer visible to the human, which later makes it impossible to give the POS machine to him—in the current example this requires that all books be visible to him. Consequently, the GTP backtracks and finds a slightly different way to place the white book (f) so that both books are then visible.

Figure 5 (a) shows that at the start, the grey book (Grey) and the white book (White) are not visible (Vis) nor reachable (Reach) to the human, as they both require an effort (E) of either 3 or 4, instead of 1. Figure 5 (d) shows how after backtracking and moving the white book, the grey one becomes visible once again. Note that something is deemed visible to a human based on a threshold on the percentage of pixels visible, using the approach presented in [16]. Our threshold here is that at least 50% of an object should be visible to the human for the fact to hold.

## V. EXPERIMENTAL RESULTS

To analyse the runtime performance of the combined HTN-GTP system we implemented and ran the domain depicted in Figure 2. Specifically, we implemented most HTN and GTP tasks in the domain except for NAVTO($Obj$) (which requires further work), and we modified the HTN domain slightly by replacing PUTON($O_1, O_2$) with MAKEACC($O, H$), and grouping PICK($O$) together with the 4 other GTP tasks for reasons explained in Section III.

We ran the experiments on the PR2, which has two quad-core i7 Xeon processors (8 cores), 24 GB of memory, and a 500 GB hard drive.[5] We only analysed the performance of the HTN-GTP system with HTN backtracking alone—we did not exploit geometric backtracking; we intend to use these results as a baseline to compare against an extended HTN-GTP system being developed, which interleaves backtracking as shown in Figure 3. We ran the experiment 100 times, where each run started by calling the MANAGEORDER($M$) task in Figure 2. The initial state of the GTP was similar to that in Figure 4 except that there was also a POS machine. We kept the GTP initial state fixed: automatically generating initial states with different positions and orientations for books is left as future work. We do note, however, that manual adjustments to the GTP initial state did not seem to change the experimental results. The HTN initial state was such that the member had to always pay by credit card, which forced method $m_7$ to be selected—the one that relies on the GTP.[6]

[4]Note that by deriving and using geometric states as shared literals we need to address the ramifications of geometric backtracking on already pursued symbolic actions. Due to space limits we leave our solution for another paper.

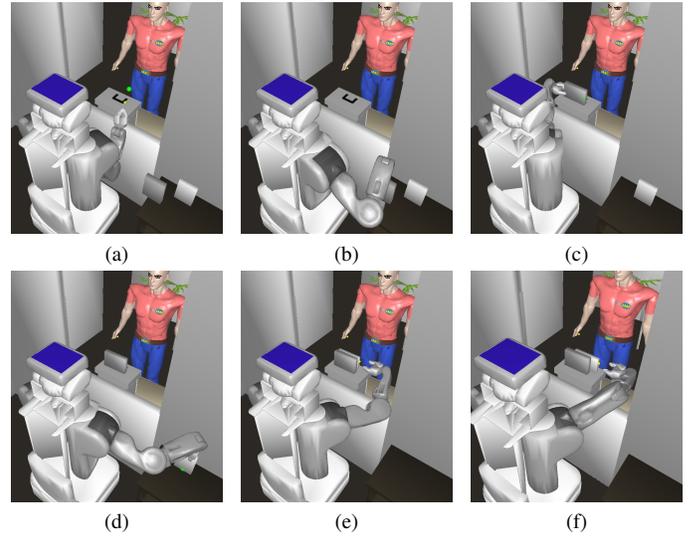

(a)    (b)    (c)

(d)    (e)    (f)

Fig. 4: A Move3D scenario for the domain in Figure 2: making two books accessible by backtracking.

| For HUMAN: | For HUMAN: |
|---|---|
| Grey is Vis with E=3, Reach with E=4 | Grey is Vis with E=1, Reach with E=1 |
| White is Vis with E=4, Reach with E=4 | White is Vis with E=4, Reach with E=4 |
| For PR2: | For PR2: |
| Grey is Vis with E=1, Reach with E=1 | Grey is Vis with E=1, Reach with E=1 |
| White is Vis with E=1, Reach with E=1 | White is Vis with E=1, Reach with E=1 |
| (a) Shared literals for Figure 4(a). | (b) Shared literals for Figure 4(d). |

| For HUMAN: | For HUMAN: |
|---|---|
| Grey is Vis with E=2, Reach with E=1 | Grey is Vis with E=1, Reach with E=1 |
| White is Vis with E=1, Reach with E=1 | White is Vis with E=1, Reach with E=1 |
| For PR2: | For PR2: |
| Grey is Vis with E=1, Reach with E=1 | Grey is Vis with E=1, Reach with E=1 |
| White is Vis with E=1, Reach with E=1 | White is Vis with E=1, Reach with E=1 |
| (c) Shared literals for Figure 4(e). | (d) Shared literals for Figure 4(f). |

Fig. 5: Symbolic facts computed for Figure 4.

The results are summarised in Table II. We note that HTN planning (alone) took negligible time. Observe that PUTAWAY($O, H$) was the most "difficult" task to plan: the GTP could not find a solution 52% of the time, and it took 51 seconds on average, making it also the most computationally expensive task. On the contrary, the GTP always found a solution for GIVE$(O, H)$, which was also the least computationally expensive task.

[5]Running the HTN-GTP system on the PR2 allowed using real object and human locations for constructing the initial states for planning, and also directly executing plans found.

[6]We invite the reader to view a video of a single run of the experiment at www.tinyurl.com/pr2-exp; it has been edited for easier viewing.

| ACTION | TIME | SUCC % | PTS | GRASPS | ORTS | CALLS |
|---|---|---|---|---|---|---|
| MAKEACC | 18.8 | 99.5 | 11 | 93.3 | 644.7 | 16.2 |
| SHOW | 13.3 | 93 | 1 | 20.3 | 406.6 | 7.8 |
| GIVE | 10.1 | 100 | 1 | 4.2 | 94 | 7.9 |
| PUTAWAY | 51 | 48 | 13 | 110.3 | 569.2 | 17.9 |

TABLE II: The table summarises, for each GTP task: the time taken to find a solution (time); what percentage of its planning attempts were successful (succ %); the total number of points tested in 3D space (pts); the total number of grasps tried (grasps); the total number of orientations tried (orts); and the total number of times the trajectory planner was called (calls).

Although, intuitively, it should be possible to find solutions less often for GIVE$(O, H)$ than for SHOW$(O, H)$, this was not the case because GIVE$(O, H)$ is followed by SHOW$(O, H)$ in our HTN domain: hence if there is no solution for SHOW$(O, H)$ then GIVE$(O, H)$ will not be attempted, and if SHOW$(O, H)$ does have a solution then it is quite likely that GIVE$(O, H)$ will also have one, as they both rely on the ability to make an object visible.[7]

Interestingly, MAKEACC$(O, H)$ and PUTAWAY$(O, H)$ have similar results in their respective columns in the table, but the latter takes much longer to plan. This is due to different thresholds on visibility: for an object to be deemed accessible it is sufficient if (along with reachability requirements) at least 50% of the object's pixels are visible to the person, whereas for an object to be considered successfully put away, we have set that value to 0%. Since our GTP approach is constraint hierarchy based, the computation time is greatly affected by what stage in the planning process the planner fails. As pixel based visibility computation is relatively computationally expensive, it is left for the final stages of planning. It appears that with PUTAWAY$(O, H)$, checking the feasibility of visibility constraints fails more times than with MAKEACC$(O, H)$, as in most cases objects were not completely hidden—in fact, they were sufficiently visible.

The experiments revealed that our combined HTN-GTP domain was, in some sense, "complete"—the combination was almost always able to find HTN-GTP solutions for MANAGEORDER$(M)$, without frequent backtracking. There was only one HTN-GTP planning attempt that failed completely (with 8 backtracks), when the GTP could not put/hide away the POS machine—within the 60-second time limit we had set for GTP—despite all symbolic backtracking attempts. On average the symbolic planner backtracked about once (actually, 0.86 times) per run before finding a solution, with slightly under 10% of the total runs backtracking more than 3 times per run. This suggests that GTP backtracking should, perhaps, be done sparingly—rather than every time a GTP task being pursued has no solution—by relying more on HTN backtracking, albeit at the expense of completeness. Another trade-off between completeness and efficiency is to significantly reduce the grasps and orientations tested by GTP tasks like MAKEACC$(O, M)$ to a few "good" ones that generally work well; the number of pixels tested when determining object visibility could also be reduced in a similar way. We believe that these improvements will make the combined system more practical for real-world applications.

## VI. CONCLUSION

We have presented an approach to combining HTN and geometric task planning, which allows more sophisticated reasoning than possible with standard trajectory planning. The combination makes way for rich backtracking at multiple levels, and also interleaved backtracking. We showed how the two planners could be interfaced, and gave insights into a methodology for developing HTN-GTP domains. Our prototype implementation is able to do basic GTP backtracking, and to compute and share symbolic facts, which the HTN developer can use in preconditions. Finally, we presented experimental results that we intend to use as a benchmark to test future extensions, and for developing heuristics for the GTP.

---

[7] While GIVE$(O, H)$ also needs $O$ to be reachable to $H$ in free space, in our scenario this is easy to achieve.